# On Separation Criterion and Recovery Algorithm for Chain Graphs*

Milan Studený[†]
Institute of Information Theory and Automation
Academy of Sciences of Czech Republic
Pod vodárenskou věží 4, 182 08 Prague, Czech Republic

## Abstract

Chain graphs (CGs) give a natural unifying point of view on Markov and Bayesian networks and enlarge the potential of graphical models for description of conditional independence structures. In the paper a direct graphical separation criterion for CGs which generalizes the d-separation criterion for Bayesian networks is introduced (recalled). It is equivalent to the classic moralization criterion for CGs and complete in the sense that for every CG there exists a probability distribution satisfying exactly independencies derivable from the CG by the separation criterion. Every class of Markov equivalent CGs can be uniquely described by a natural representative, called the largest CG. A recovery algorithm, which on basis of the (conditional) dependency model given by a CG finds the corresponding largest CG, is presented.

## 1 INTRODUCTION

Traditional graphical models for description of probabilistic conditional independence structure use either undirected graphs (UGs), named also Markov networks, or directed acyclic graphs (DAGs), known as Bayesian networks or (probabilistic) influence diagrams. In middle eighties Lauritzen and Wermuth (1984) introduced the class of *chain graphs* (CGs) which involves both UGs and DAGs, but not only them. In CGs both undirected edges, called *lines*, and directed edges, called *arrows*, are simultaneously allowed, but directed cycles are forbidden (nevertheless, undirected cycles are allowed). To establish semantics for CGs Lauritzen (1989) and Frydenberg (1990a) generalized so-called *moralization criterion* for DAGs from (Lauritzen et. al. 1990) for reading independencies from a CG. As in case of DAGs, it consists of 3 steps: restriction of the CG to a certain set of nodes, transforming it properly to an UG (called the *moral graph*), and using the separation criterion for UGs with respect to the moral graph. Moreover, they conjectured that for every CG there exists a probability distribution which exhibits exactly those independency statements which can be read from the graph according to the moralization criterion.

On basis of the moralization criterion Frydenberg (1990a) characterized Markov equivalent CGs (that is CGs producing the same dependency model) in graphical terms, namely as CGs having the same underlying graph and the same occurencies of complexes. Note that it generalizes an analogous characterization of Markov equivalence for DAGs from (Verma and Pearl 1991). Moreover, Frydenberg (1990a) showed that every class of Markov equivalent CGs has a CG with the greatest number of lines (or dually with the least number of arrows). Thus, every class of Markov equivalent CGs can be naturally represented by this distinguished CG, called the *largest CG*.

Several recent works suggest that CGs attract attention of researchers. Whittaker (1990) is his book gave several examples of use of CGs, Cox and Wermuth (1993) stimulated a discussion on them. Andersson, Madigan and Perlman (1995) used special CGs, called *essential graphs*, to represent uniquely classes of Markov equivalent DAGs and characterized them in graphical terms. Note for explanation that the essential graph for a class of Markov equivalent DAGs does not coincide in general with the largest CG for the corresponding class of Markov equivalent CGs. In UAI'95 conference two papers were dealing with CGs: (Buntine 1995) gave an equivalent definition of CGs as hierarchical combination of Markov and Bayesian networks and (Meek 1995) generalized an algorithm from (Verma and Pearl 1992) which on basis of the dependency model corresponding to a DAG finds the above mentioned essential graph.

In (Bouckaert Studený 1995) we have introduced a graphical *separation criterion*, called *c-separation* ('c'

*This work was supported by the grant of Grant Agency of Czech Republic n. 201/94/0471 and CEC n. CIPA3511CT930053.
[†]E-mail: studeny@utia.cas.cz

stands for 'chain'), for reading independencies from a CG. Unlike the moralization criterion it tests directly trails in the original CG (i.e. the graph is not transformed to an UG) and in this sense it generalizes the well-known d-separation criterion for DAGs from (Pearl 1988). Moreover, we have shown in (Studený Bouckaert 1996) that the new criterion is equivalent to the moralization criterion and have used it to confirm the above mentioned Lauritzen's and Frydenberg's conjecture, that is to prove the completeness of both criteria. This generalizes analogous results for UGs and DAGs from (Frydenberg 1990b) (Geiger Pearl 1993) and (Geiger Pearl 1990). In the first part of this paper these new concepts and results are recalled.

In the second part so-called *recovery algorithm* is presented. On basis of a dependency model, which is supposed to correspond to an unknown CG, it finds the largest CG of the corresponding class of Markov equivalent CGs. Like the procedures from (Verma Pearl 1992) (Spirtes et. al. 1993) (Meek 1995) it has two stages. The first one is strongly related to Frydenberg's (1990a) characterization of Markov equivalence of CGs. On basis of special 'elementary' statements obtained from the dependency model one identifies the edges of the underlying graph and the occurencies of complexes and forms so-called *pattern* of the equivalence class. It is a special graph, having the required underlying graph and only arrows produced by the complexes (the other edges are lines). However, this graph is not a CG in general (it may have oriented cycles) and some of its lines has to be directed to obtain the corresponding largest CG. This is made in the second stage by repeated application of certain orientation rules.

The structure of the paper is as follows. In the next section basic definitions are recalled and essential results reviewed. In the third section the concept of c-separation is introduced and illustrated by an example. The corresponding results are formulated, too. The pattern of the corresponding equivalence class is constructed in the fourth section. The fifth section describes the method of obtaining the largest CG on basis of the pattern. In Conclusions several remarks on obtained results are given. The Appendix contains the proof of the correctness of the algorithm from the fourth section.

## 2 BASIC CONCEPTS

### 2.1 GRAPHS

A *hybrid graph* $G$ over a nonempty finite set of *nodes* $N$ is specified by a set of two-element subsets of $N$, called *edges*, where every edge $\{u,v\}$ is either a *line* (= undirected edge), denoted by $u - v$, or an *arrow* (= directed edge) from $u$ to $v$, denoted by $u \to v$, or an arrow from $v$ to $u$, denoted by $u \leftarrow v$. An *undirected graph* (UG) is a graph containing lines only, a *directed graph* is a graph containing arrows only. The *underlying graph* of $G$ is obtained from $G$ by changing all edges of $G$ into lines. An *induced subgraph* of $G$ on a nonempty set $T \subset N$, denoted by $G_T$, is the graph over $T$ which has exactly those edges of $G$ which are subsets of $T$. *Connectivity components* of $G$ are obtained by removing all arrows of $G$ and taking the connectivity components of the remaining undirected graph. A *route* in $G$ is a sequence of its nodes $v_1, \ldots, v_k$, $k \geq 1$ such that $\{v_i, v_{i+1}\}$ is an edge of $G$ for every $i = 1, \ldots, k-1$. It is called a *path*, if it consists of distinct nodes. It is called a *pseudocycle* if $v_1 = v_k$, and a *cycle*, if moreover $v_1, \ldots, v_{k-1}$ are distinct. A (pseudo)cycle is *directed*, if $v_i \to v_{i+1}$ or $v_i - v_{i+1}$ for $i = 1, \ldots, k-1$, and surely $v_j \to v_{j+1}$ for at least one $j \in \{1, \ldots, k-1\}$. A *directed acyclic graph* (DAG) is a directed graph without directed cycles.

A *complex* in $G$ is its special induced subgraph, namely a path $v_1, \ldots, v_k$, $k \geq 3$, such that $v_1 \to v_2$, $v_i - v_{i+1}$ for $i = 2, \ldots, k-2$, $v_{k-1} \leftarrow v_k$ in $G$, and no additional edges between nodes of $\{v_1, \ldots, v_k\}$ exist in $G$. The nodes $v_1$ and $v_k$ are called the *parents* of the complex, the set $\{v_2, \ldots, v_{k-1}\}$ the *region* of the complex and the number $k-2$ is the *degree* of the complex. Note that the concept of complex is equivalent to Frydenberg's (1990a) notion of 'minimal complex'. I decided to simplify the terminology: I believe that 'nonminimal complexes' have no reasonable use.

A *chain* for $G$ is a partition of $N$ into ordered disjoint (nonempty) subsets $B_1, \ldots, B_n$, $n \geq 1$ called *blocks* such that, if $\{u,v\}$ is an edge with $u,v \in B_i$ then $u - v$, and if $\{u,v\}$ is an edge with $u \in B_i, v \in B_j, i < j$ then $u \to v$. The original definition of CG (which also explains the terminology) is the following one.

DEFINITION 2.1 A *chain graph* (CG) is a hybrid graph which admits a chain.

However, there are several equivalent definitions of CG, whose verification is left to the reader. They imply that CGs involve both UGs and DAGs.

LEMMA 2.1 *The following conditions are equivalent for a hybrid graph* $G$.

- $G$ *is a chain graph*,
- $G$ *has no directed pseudocycles*,
- $G$ *has no directed cycles*,
- *the set of connectivity components of* $G$ *can be ordered to form a chain.*

Note that one CG may admit several chains, but every block of a chain is a union of connectivity components of the graph. Thus, chains made of connectivity components cannot be refined.

Having nodes $u, v$ with $u \to v$ in $G$, $u$ is called a *parent* of $v$ and $v$ a *child* of $u$. In case $u - v$ they are *siblings*.

The set of parents, resp. children of a node $u$ in $G$ is denoted by $pa_G(u)$, resp. $ch_G(u)$. The *boundary* of $u$, denoted by $bd_G(u)$, is the set of parents and siblings of $u$. The symbol of the graph $G$ is omitted when it is clear from context. A path $v_1, \ldots, v_k$, $k \geq 1$ is *descending* if $v_i \to v_{i+1}$ or $v_i - v_{i+1}$ for $1 \leq i \leq k-1$. Especially, an undirected path is considered as a descending path. If there exists a descending path from a node $u$ to a node $v$, then $u$ is an *ancestor* of $v$, or dually $v$ is a *descendant* of $u$. Having a set of nodes $A \subset N$ its *ancestral set*, denoted by $an_G(A)$, is the set of all ancestors of nodes in $A$ (it contains $A$).

## 2.2 DEPENDENCY MODELS

Supposing $N$ is a nonempty finite set of variables let us denote by $T(N)$ the class of triplets $\langle X, Y | Z \rangle$ of disjoint subsets of $N$ whose first two components $X$ and $Y$ are nonempty. A *dependency model* over a $N$ is a decomposition of $T(N)$ into two parts, namely the *independency part* and the complementary *dependency part*. Let's write $I_M \langle X, Y | Z \rangle$ if a triplet $\langle X, Y | Z \rangle$ belongs to the independency part of a dependency model $M$, otherwise write $D_M \langle X, Y | Z \rangle$.

A *probability distribution over* $N$ is specified by a collection of nonempty finite sets $\{ \mathbf{X}_i ; i \in N \}$ and by a function $P : \prod_{i \in N} \mathbf{X}_i \to [0,1]$ with $\sum \{ P(\mathbf{x}); \mathbf{x} \in \prod_{i \in N} \mathbf{X}_i \} = 1$. If $P(\mathbf{x}) > 0$ for all $\mathbf{x} \in \prod_{i \in N} \mathbf{X}_i$, then $P$ is called *strictly positive*. As I consider the concepts from Pearl's (1988) book like of (probabilistic) *conditional independence*, *graphoid* and *semigraphoid* to be well-known to UAI community, I decided to omit their definitions. Nevertheless, I recall some further basic concepts, used in formulation of results. A *probabilistic* dependency model *induced by* a probability distribution $P$ over $N$ has specified its independency part as the collection of all triplets $\langle X, Y | Z \rangle \in T(N)$ representing valid conditional independencies in $P$. Note that it is a semigraphoid and a graphoid, if induced by a strictly positive distribution (Dawid 1979). A triplet $t \in T(N)$ belongs to the *probabilistic closure* of a set $L \subset T(N)$ w.r.t. a class of probability distributions $\mathcal{P}$ over $N$, if $t$ represents a valid conditional independency in every $P \in \mathcal{P}$ such that every triplet from $L$ represents a valid conditional independency in $P$.

Supposing $G$ is a CG, its *moral graph* is obtained in two steps. First, the parents of every complex in $G$ are joined by an edge. Second, the underlying graph of the resulting graph is taken. (Frydenberg 1990a) gave another equivalent definition, namely to join the parents of every connectivity component of $G$ which are not joined, and then to 'forget' the orientations. Note that in (Studený Bouckaert 1996) we have kept the original type of edges in the moral graph and used special 'virtual' edges to join the parents of complexes. It was very convenient in the context of that paper, but here it is immaterial.

A triplet $\langle X, Y | Z \rangle \in T(N)$ is *represented* in a CG $G$ according to the *moralization criterion* if every path in the moral graph of $G_{an(X \cup Y \cup Z)}$ from a node of $X$ to a node of $Y$ meets $Z$. Thus, the moralization criterion taken from (Lauritzen 1989) or (Frydenberg 1990a) has 3 steps. First, to take the induced subgraph of $G$ on the corresponding ancestral set $an_G(X \cup Y \cup Z)$. Second, to find the moral graph of the induced subgraph. Third, to apply the classic separation criterion for UGs to that moral graph.

The dependency model *induced by* a CG $G$ has specified its independency part as the class of triplets represented in $G$ according to the moralization criterion. It is not difficult to show that it is a graphoid (Bouckaert Studený 1995). A probability distribution $P$ over $N$ is *Markovian* w.r.t. a CG $G$ if every triplet represented in $G$ (according to the moralization criterion) belongs to the probabilistic independency model induced by $P$. Two graphs $G$ and $H$ over $N$ are *Markov equivalent* if their classes of Markovian distributions coincide. (Frydenberg 1990a) gave the following characterization which generalizes an analogous result for DAGs from (Verma Pearl 1991).

THEOREM 2.1 *Two CGs are Markov equivalent iff they have the same underlying graph and complexes.*

Supposing $G$ and $H$ are CGs over the same set of variables with the same underlying graph, we say that $G$ is *larger* than $H$, denoted by $H \prec G$, if every arrow of $G$ is an arrow in $H$ with the same orientation. Note that (Frydenberg 1990a) defined the relation 'larger' for every couple of CGs and I use only a restricted definition here. The following theorem reformulates a little bit further Frydenberg's (1990a) result.

THEOREM 2.2 *For every CG $G$ there exists a Markov equivalent CG $G_\infty$, such that $H \prec G_\infty$ for every CG $H$ which is Markov equivalent to $G$.*

DEFINITION 2.2 The graph $G_\infty$ from the previous theorem is called the *largest* CG corresponding to $G$.

## 3  SEPARATION CRITERION

In (Bouckaert Studený 1995) we have introduced a direct separation criterion for CGs, which generalizes the concept of d-separation for DAGs from (Pearl 1988). The *c-separation* (chain separation) criterion exhibits two main differences from the case of DAGs. First, one has to consider a wider class of routes (not only paths consisting of distinct nodes). Second, the blocking of the route is not defined for nodes of the route, but for its maximal undirected subroutes, called *sections*.

Thus, every route decomposes uniquely into its sections, and sections can be classified according to the orientations (resp. the existence) of arrows which delimit them. A *head-to-head* section has two incoming arrows, a *head-to-tail* section one incoming arrow only, and *tail-to-tail* section no incoming arrow. A node $u$ of a section $\sigma$ of a route $\rho$ is a *tail-terminal* node of $\sigma$

512  Studeny

w.r.t. $\rho$ if $u$ is the last node of $\sigma$ ($=$ a terminal node of $\sigma$) and moreover, either $u$ is also the last node of $\rho$ (more precisely $\sigma$ is not limited at $u$ by an arrow of $\rho$ because of $\rho$ ends at $u$ already) or the corresponding arrow of $\rho$ delimiting $\sigma$ at $u$ emanates from $u$ (i.e. it is an outgoing arrow from $u$). A *trail* in a CG is such a route that no arrow appears twice (or more times) in it, and whose every section consists of distinct nodes. A *slide* from a node $v_1$ to a node $v_k$ is a path $v_1, \ldots, v_k$, $k \geq 2$ such that $v_1 \to v_2$, and $v_i - v_{i+1}$ for all $i = 2, \ldots, k-1$. A section $\sigma$ of a trail in a CG is *blocked* by a set of nodes $Z \subset N$ if one of the following two cases occurs:

**either** $\sigma$ is a head-to-head section and has no descendant in $Z$,

**or** $\sigma$ is not a head-to-head section, $\sigma$ meets $Z$, and moreover for some (at least one) its tail-terminal node $u$, every slide to $u$ in $G$ meets $Z$.

A trail is *c-separated* by a set $Z$ if at least one of its sections is blocked by $Z$, otherwise it is called *active* w.r.t. $Z$. The reader can verify that in case of DAGs it collapses to d-separation. A triplet $\langle X, Y | Z \rangle \in T(N)$ is *represented* in a CG according to the *separation criterion* if every trail from a node of $X$ to a node of $Y$ is c-separated by $Z$. We have proved in (Studený Bouckaert 1996, Consequence 4.1):

**THEOREM 3.1** *The moralization criterion and the separation criterion for CGs are equivalent.*

The following example shows that one cannot limit oneself to paths in the c-separation criterion. On the other hand, one should realize that only finitely many trails exist between two different nodes in every CG.

EXAMPLE 3.1 Let us consider the CG in the figure 1 and test the triplet $\langle a, f | ceg \rangle$ by the separation criterion. The only path from $a$ to $f$ is $a \to c - d \to f$. Its tail-to-tail section $a$ is not blocked by $Z = ceg$, because of it does not meet $Z$, similarly its head-to-tail section $f$. However, its head-to-tail section $c - d$ is blocked (there is no slide to its tail-terminal node $d$ which is outside $Z$). But, c-separation is not limited to paths and one has to consider also the trail $a \to c - d - e \leftarrow b \to g \leftarrow d \to f$. It is active: both its head-to-head sections $c - d - e$ and $g$ have a descendant in $Z$ and all other sections do not meet $Z$. Thus, $\langle a, f | ceg \rangle$ is not represented in the CG according to the separation criterion. The reader can obtain the same conclusion by the moralization criterion.

Supposing $G$ be a CG over a $N$ and $B_1, \ldots, B_n$ a chain for it, the associated *input list* $L$ is the collection of triplets $\langle u, B_1 \cup \ldots \cup B_{k(u)} \setminus bd_G(u) \cup \{u\} | bd_G(u) \rangle$, where $u \in N$ and $B_{k(u)}$ denotes the block containing $u$. Note that generalizes analogous concepts for DAGs from (Verma Pearl 1990) or UGs (Bouckaert 1995). We have proved in (Studený Bouckaert 1996, Theorems 7.1 and 7.2) the following results.

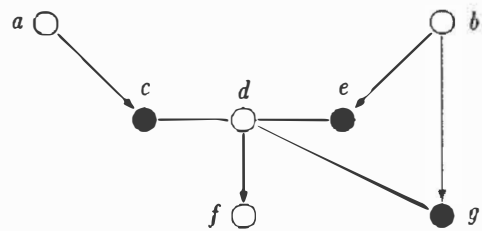

Figure 1: An illustrative example for c-separation

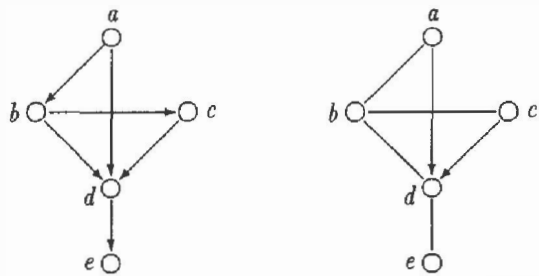

Figure 2: Example of a DAG and its pattern

**THEOREM 3.2** *Supposing $G$ is a CG and $L$ the input list associated with a chain for $G$, the following conditions are equivalent for $t \in T(N)$:*

- *$t$ is represented in $G$,*
- *$t$ belongs to the graphoid closure of $L$,*
- *$t$ belongs to the probabilistic closure of $L$ w.r.t. the class of strictly positive probability distributions.*

**THEOREM 3.3** *For every CG $G$ over $N$ there exists a strictly positive probability distribution $P$ over $N$ such that the dependency models induced by $G$ and $P$ coincide.*

## 4  PATTERN RECOVERY

The first step of the recovery algorithm is the pattern of the corresponding class of Markov equivalent CGs.

DEFINITION 4.1 Supposing $G$ is a CG, its *pattern*, denoted by $G_0$, is a hybrid graph obtained from the underlying graph of $G$ by directing all edges which are arrows in a complex of $G$ (with the same orientation).

It follows from Theorem 2.1 that two CGs are Markov equivalent iff they have the same pattern. However, the pattern may not be a CG, as the following example shows.

EXAMPLE 4.1 To illustrate the concept of pattern, let us consider the DAG $G$ in the left picture of the figure

2. It has only one complex $a \to d \leftarrow c$. The corresponding pattern is in the right picture of the figure.

To reconstruct the pattern of a CG from its induced dependency model the following notation is suitable.

DEFINITION 4.2 Let $M$ be a dependency model over $N$ and $u, v, w \in N$ are distinct. The symbol $D_M\langle u, v|-\rangle$ will be used to replace an entire collection of statements, namely $D_M\langle u, v|Z\rangle$ for all $Z \subset N \setminus \{u, v\}$. Similarly, $D_M\langle u, v| + w\rangle$ will substitute $D_M\langle u, v|Z\rangle$ for all $Z \subset N \setminus \{u, v\}$ with $w \in Z$.

The algorithm presented here produces a sequence of hybrid graphs $H_i$ with the same underlying graph as $G$, such that $H_i$ has all complexes of $G$ of degree at most $i$.

PATTERN RECOVERY ALGORITHM Let $M$ be the dependency model induced by an (unknown) CG $G$ over $N$.

(i) The starting iteration is an undirected graph $H_0$ over $N$ defined by the following rule: $u - v$ in $H_0$ iff $D_M\langle u, v|-\rangle$.

(ii) The next iteration $H_1$ is made by directing some lines of $H_0$, namely for every three distinct nodes $u, v, w$ of $N$ such that $u - w - v$ in $H_0$, $\neg(u - v)$ in $H_0$, and $D_M\langle u, v| + w\rangle$ one has $u \to w \leftarrow v$ in $H_1$. Other lines of $H_0$ remain lines in $H_1$.

(iii) For $l = 2, \ldots,$ card $N - 2$ the iteration $H_l$ is made from $H_{l-1}$ by possible directing of some lines of $H_{l-1}$. Namely, in every situation when some sequence of distinct nodes $w_1, \ldots, w_{l+2}$ exists such that $w_1 \to w_2$ or $w_1 - w_2$ in $H_{l-1}$, $w_{l+1} \leftarrow w_{l+2}$ or $w_{l+1} - w_{l+2}$ in $H_{l-1}$, $w_i - w_{i+1}$ in $H_{l-1}$ for $i = 2, \ldots, l$, no other edge exists in $H_{l-1}$ among $\{w_1, \ldots, w_{l+2}\}$, $D_M\langle w_1, w_{l+2}| + w_2\rangle$, and $D_M\langle w_1, w_{l+2}| + w_{l+1}\rangle$, one has $w_1 \to w_2$ and $w_{l+1} \leftarrow w_{l+2}$ in $H_l$. Note that these edges may be possibly directed already in $H_{l-1}$. All other edges of $H_l$ keep their type and orientation from $H_{l-1}$.

The following result is proved in Appendix.

THEOREM 4.1 *The last iteration of the previous algorithm is nothing but the pattern of the considered (unknown) CG $G$.*

## 5 LARGEST CG RECOVERY

The obtained pattern $G_0$ of the considered class of Markov equivalent CGs is to be changed into the corresponding largest CG $G_\infty$. Iterations of the presented algorithm are not mere hybrid graphs, but hybrid graphs, some lines of which have 'forbidden' potential orientations in $G_\infty$. Let us write $\neg\{u \leftarrow v\}_\infty$ to denote that a line $u - v$ has forbidden the orientation $u \leftarrow v$ in $G_\infty$. A *feasible semislide* from a node $w_1$ to a node $w_k$ in such a graph $H$ is a route $w_1, \ldots, w_k$, $k \geq 2$ where $w_1 \to w_2$ in $H$, and for $i = 2, \ldots, k-1$ either $w_i \to w_{i+1}$ in $H$ or $[w_i - w_{i+1}$ and $\neg\{w_i \leftarrow w_{i+1}\}_\infty$ in $H]$.

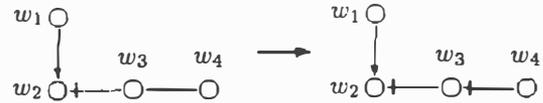

Figure 3: Transitivity principle

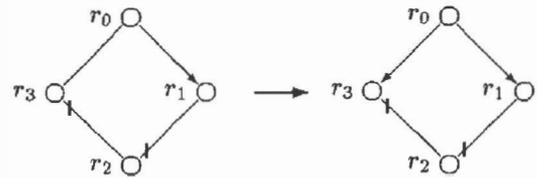

Figure 4: Necessity principle

LARGEST CHAIN GRAPH RECOVERY ALGORITHM Let $G_0$ be a hybrid graph over $N$ which is the pattern of a class of Markov equivalent CGs. The algorithm alternates 'orientation bans' by a *transitivity principle* and 'edge directing' either by a *necessity principle* or by a *doublecycle principle*. However, orientation bans have priority.

Transitivity principle (see the figure 3) means that whenever $w_1, \ldots, w_k$ is a feasible semislide in $G_l$ ($l \geq 0$), $w_k - w_{k+1}$ in $G_l$, and there is no edge in $G_l$ between $w_{k+1}$ and $\{w_1, \ldots, w_{k-1}\}$, then $\neg\{w_k \leftarrow w_{k+1}\}_\infty$ in $G_{l+1}$, i.e. $w_1, \ldots, w_{k+1}$ is a feasible semislide in $G_{l+1}$.

Note that the transitivity principle ensures that all lines in complexes in $G$ will have forbidden both orientations in $G_\infty$.

Necessity principle (see the figure 4) means that whenever $r_0, \ldots r_k$, $k \geq 2$ is a pseudocycle in $G_l$ ($r_{k+1} = r_0$) such that $r_0 \to r_1$ in $G_l$, $r_j - r_{j+1}$ in $G_l$ for some $1 \leq j \leq k$, and for every $i \in \{1, \ldots, k\} \setminus \{j\}$ either $r_i \to r_{i+1}$ in $G_l$ or $[r_i - r_{i+1}$ and $\neg\{r_i \leftarrow r_{i+1}\}_\infty$ in $G_l]$, then the line $r_j - r_{j+1}$ is changed into the arrow $r_j \leftarrow r_{j+1}$ in $G_{l+1}$.

Doublecycle principle (see the figure 5) assumes that $r_0, \ldots r_k$, $k \geq 2$ is a pseudocycle in $G_l$ such that $r_0, \ldots r_{k-1}$ is a feasible semislide in $G_l$, and $r_{k-1} - r_k$, $r_k - r_{k+1} = r_0$ in $G_l$. Moreover it supposes that $s_0, \ldots, s_m$, $m \geq 1$ is a feasible semislide to $r_1 = s_m$ such that $s_0 \neq r_0$ and there exists $0 \leq n \leq m-1$ such that $\{r_k, s_n\}$ is an edge in $G_l$, but there is no edge in $G_l$ between $r_0$ and $\{s_0, \ldots, s_n\}$. Then the line $r_{k-1} - r_k$ in $G_l$ is changed into the arrow $r_{k-1} \leftarrow r_k$ in $G_{l+1}$.

Note that the edge $\{r_{k-1}, r_k\}$ which is directed by the previous principle belongs to two pseudocycles, namely

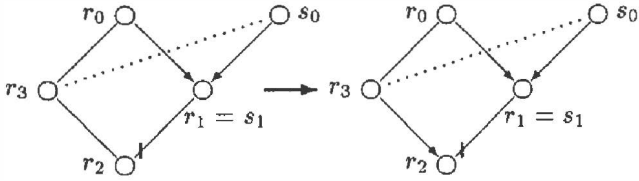

Figure 5: Doublecycle principle

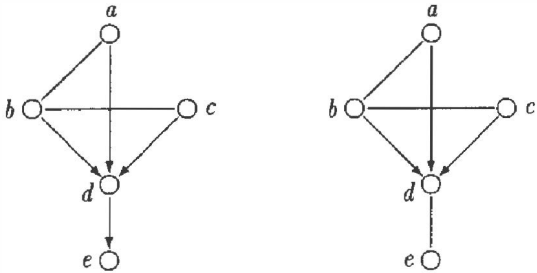

Figure 6: An essential graph and largest CG

to $r_0, \ldots r_k$ and $s_n, \ldots, s_m = r_1, \ldots r_k$. This motivated the terminology.

THEOREM 5.1 *The result of the described algorithm is the largest CG of the corresponding class of Markov equivalent CGs.*

The proof of the Theorem 5.1 is beyond the scope of a conference contribution. It will be given in (Studený 1996). Let us conclude the section by an example.

EXAMPLE 5.1 Let us consider the DAG from figure 2. The corresponding essential graph, called the 'completed pattern' in (Verma Pearl 1991), is given in the left picture of figure 6. The largest CG of the corresponding class of Markov equivalent CGs is in the right picture of the same figure.

## 6 CONCLUSIONS

Several remarks conclude the contribution. The first remark concerns the significance of the concept of largest CG. Markov networks have one big advantage: different UGs yield different dependency models. Bayesian networks have not such pleasant property: two different DAGs may represent the same dependency model, that is to be Markov equivalent. Moreover, the class of Markov equivalent DAGs has no natural representative and one has to represent the class by a pattern or by an essential graph. However, then the problem arises whether such type of representation allows to identify back the corresponding dependency model. As patterns and essential graphs are not DAGs in general one cannot use the criteria for DAGs to obtain the dependency model. However, the concept of largest CG provides a reasonable solution even in case of Bayesian networks. One can represent the class of Markov equivalent DAGs by the largest CG of the corresponding class of Markov equivalent CGs (which is, of course, wider, but represents the same dependency model). As the largest CG is a real member of the class of Markov equivalent CGs, one can identify back the corresponding dependency model by some criterion for CGs, for example c-separation. In fact, the concept of essential graph also provides a solution of the mentioned problem because of it also belongs to the class of CGs which are Markov equivalent to the considered DAG - see (Andersson et. al. 1995). I think that this result is a good argument why one should not be fixed strictly only to DAG-models: a wider perspective of CGs may solve simply problems specific for Bayesian networks. Thus, both the essential graphs and the largest CGs solve the above mentioned problem for DAGs, but the use of the largest CG is not limited to DAG-models.

The second remark concerns the significance of the separation criterion for CGs. It is more intuitive in nature than the moralization criterion and this gave the way to the proof of Theorems 3.2 and 3.3 which justify completely the use of CGs for the description of probabilistic dependency models. Note that Theorem 3.2 can be interpreted as the statement that graphoid properties are complete for input lists - in this respect it generalizes an analogous result for DAGs from (Verma Pearl 1990).

The third remark concerns the pattern recovery algorithm. It has an important feature: it depends only on predicates $\langle u, v | - \rangle$ and $\langle u, v | + w \rangle$ introduced in Definition 4.2. Especially, two CG-models which coincide on these predicates must be equal! The number of such predicates is polynomial in number of variables unlike the exponential number of triplets in a general dependecy model. This may give a more precise estimate of the number of CG-models or DAG-models. Perhaps a representation of DAG-models in terms of these predicates would be more effective.

The fourth remark concerns the largest CG recovery algorithm. The orientation rules (principles) are formulated in very general form - for semislides and pseudocyles. The reason is that I succeeded to prove their completeness just in this general form. However, I conjecture that they are complete also when they are formulated in a 'narrow' sense: that is the transitivity principle for feasible slides, the necessity and the doublecycle principles for minimal cycles (i.e. without a chord) and feasible slides. In all examples I studied, such restricted formulation was sufficient, but technical complications hindered me in the proof of such a stronger result. The question of computational complexity of the algorithm is an interesting question which can be a topic of further research.

Note that in this paper attention was restricted to probability distributions on finite sets just for simplic-ity and clarity. Analogous concepts can be considered also in continous case.

### Acknowledgements

I would like to express my gratitude to all three anonymous reviewers who gave me several comments which helped me to improve the paper. I also apology that for page limitation of the paper I did not respond to all their interesting suggestions for discussion.

## 7 APPENDIX

**LEMMA 7.1** *Let $G$ be a $CG$ over $N$, $M$ the dependency model induced by $G$, $u, v \in N$ distinct such that $\{u, v\}$ is not an edge of $G$. Then $I_M\langle u, v \mid bd_G(u) \cup bd_G(v)\rangle$.*

**Proof:** Let us apply the moralization criterion to $\langle u, v | T\rangle$ where $T = bd_G(u) \cup bd_G(v)$. Evidently $an_G(\{u, v\} \cup T) = an_G(\{u, v\})$ and one is to consider the induced subgraph $H = G_{an(\{u,v\})}$. Let us verify by contradiction that either $ch_H(u)$ or $ch_H(v)$ is empty. Indeed, if $u \to t$ in $H$ for some $t$, then owing to $t \in an_G(\{u, v\})$ there exists a descending path in $G$ from $t$ to $\{u, v\}$. It has to lead to $v$, as otherwise a directed cycle in $G$ exists - see Lemma 2.1. Similarly, $v \to s$ in $H$ for some $s$ implies that there is a descending path from $s$ to $u$. Thus, if both $u \to t$ and $v \to s$, then $u \to t \ldots v \to s \ldots u$ is a directed pseudocycle in $G$ what contradicts Lemma 2.1. Therefore, one can suppose without loss of generality that $ch_H(u) = \emptyset$. This implies that no edge to $u$ is added when the moral graph of $H$ is made and $u$ has $bd_G(u)$ as the set of its neighbours in the moral graph. Thus, $T$ meets every path between $u$ and $v$ in $H$. □

**LEMMA 7.2** *Let $G$ be a CG over $N$, $M$ the dependency model induced by $G$, $u, v \in N$ distinct. Then $\{u, v\}$ is an edge of $G$ iff $D_M \langle u, v|- \rangle$.*

**Proof:** For necessity use the separation criterion: the edge $\{u, v\}$ is an active path w.r.t. any $Z \subset N \setminus \{u, v\}$. The sufficiency follows directly from Lemma 7.1. □

**LEMMA 7.3** *Let $G$ be a CG over $N$, $M$ the dependency model induced by $G$. Suppose that $u, v, w \in N$ are distinct nodes of $N$ such that $\{u, w\}, \{v, w\}$ are edges of $G$ and $\{u, v\}$ is not an edge of $G$. Then $u \to w \leftarrow v$ is a complex in $G$ iff $D_M \langle u, v| + w \rangle$.*

**Proof:** For necessity use the separation criterion: the path $u \to w \leftarrow v$ is an active path w.r.t. any $Z \subset N \setminus \{u, v\}$ containing $w$. For sufficiency one can suppose by contradiction that the induced subgraph on $\{u, w, v\}$ is not a complex in $G$. Then $w \in bd_G(u) \cup bd_G(v)$ and Lemma 7.1 leads to contradiction. □

**LEMMA 7.4** *Let $G$ be a CG over $N$, $M$ the dependency model induced by $G$. Suppose that $w_1, \ldots, w_k$, $k \geq 4$ is a sequence of distinct nodes such that*

- *$\{w_i, w_{i+1}\}$ is an edge of $G$ for $i = 1, \ldots, k-1$,*
- *no other edge exists in $G$ among $\{w_1, \ldots, w_k\}$,*
- *$w_i \to w_{i+1} - \ldots - w_{j-1} \leftarrow w_j$ is not a complex in $G$ for $1 \leq i \leq j \leq k$, $j - i < k - 1$.*

*Then $w_1 \to w_2 - \ldots - w_{k-1} \leftarrow w_k$ is a complex in $G$ iff $[D_M \langle w_1, w_k| + w_2 \rangle \& D_M \langle w_1, w_k| + w_{k-1} \rangle]$.*

**Proof:** For necessity use again the separation criterion: the path $w_1 \to w_2 - \ldots - w_{k-1} \leftarrow w_k$ is an active path w.r.t. any $Z \subset N \setminus \{w_1, w_k\}$ containing either $w_2$ or $w_{k-1}$. To show the sufficiency let us verify first by contradiction that $w_1 \to w_2$. Indeed, otherwise $w_2 \in bd_G(w_1)$ and one can use Lemma 7.1 for $u = w_1$, $v = w_k$ to get contradiction with $D_M \langle w_1, w_k| + w_2 \rangle$. Similarly, $D_M \langle w_1, w_k| + w_{k-1} \rangle$ implies $w_{k-1} \leftarrow w_k$ in $G$. Supposing that $w_{j-1} \leftarrow w_j$ in $G$ for some $3 \leq j \leq k-1$ let us consider the minimal such $j$ and find maximal $1 \leq i \leq j-1$ with $w_i \to w_{i+1}$ in $G$. Then $w_i \to w_{i+1} - \ldots - w_{j-1} \leftarrow w_j$ is a complex in $G$ what contradicts the assumption. Thus, no such $j$ exists. Similarly, no $2 \leq i \leq k-2$ with $w_i \to w_{i+1}$ in $G$ exists and $w_r - w_{r+1}$ for $r = 1, \ldots, k-1$. □

**Proof of Theorem 4.1:** By Lemma 7.2 $H_0$ has the same underlying graph as $G$, and hence every $H_i$ has the same underlying graph as $G$. Let us verify by induction on $l = 1, \ldots, \text{card } N - 2$ the following two conditions.

(a) $u \to w$ in $H_l$ implies $u \to w$ in $G_0$,

(b) every complex in $G$ of degree at most $l$ is also a complex in $H_l$.

To verify (a) for $H_1$ realize that $u \to w$ in $H_1$ implies the existence of a third node $v$ with $u - w - v$ in $H_0$, $\neg(u - v)$ in $H_0$, and $D_M \langle u, v| + w \rangle$. Hence $\{u, w\}$, $\{v, w\}$ are edges of $G$ while $\{u, v\}$ is not an edge of $G$, and by Lemma 7.3 (sufficiency) $u \to w \leftarrow v$ is a complex in $G$, which says $u \to w$ in $G_0$.

To verify (b) for $H_1$ suppose that $u \to w \leftarrow v$ is a complex in $G$ and by Lemma 7.3 (necessity) derive $D_M \langle u, v| + w \rangle$. Moreover, evidently $u - w - v$ in $H_0$, $\{u, v\}$ is not an edge of $H_0$ and thus, by construction of $H_1$, $u \to w \leftarrow v$ is a complex in $H_1$.

Supposing (a),(b) hold for $H_{l-1}$, $l \geq 2$ let us verify (a) for $H_l$. If $u \to w$ in $H_l$, then either $u \to w$ in $H_{l-1}$ and one can use the induction assumption, or, by the construction of $H_l$, there exists $w_1, \ldots, w_{l+2}$, $w_1 = u$, $w_2 = w$ such that the collection of conditions from the item **(iii)** of the algorithm is satisfied. As $H_{l-1}$ has the same underlying graph as $G$ the first two conditions of Lemma 7.4 for $k = l + 2$ are fulfilled. The third condition of Lemma 7.4 then follows from the condition (b) for $H_{l-1}$. Thus, by Lemma 7.4 (sufficiency) $w_1 \to w_2 - \ldots - w_{l+1} \leftarrow w_{l+2}$ is a complex in $G$, which implies $u = w_1 \to w_2 = w$ in $G_0$.

Supposing (a),(b) hold for $H_{l-1}$, $l \geq 2$ and (a) holds for $H_l$, let us verify (b) for $H_l$. Let $w_1 \to w_2 - \ldots - w_{s+1} \leftarrow w_{s+2}$, $s \leq l$, be a complex in $G$. First, let us show by contradiction that $w_r - w_{r+1}$ in $H_l$ for $r = 2, \ldots, s$. Indeed, if for instance $w_r \leftarrow w_{r+1}$ in $H_l$ for some $2 \leq r \leq s$, then by (a) for $H_l$, $w_r \leftarrow w_{r+1}$ in $G_0$ and therefore in $G$, what contradicts the assumption. Similar contradiction can be obtained if $w_r \to w_{r+1}$ in $H_l$ for some $2 \leq r \leq s$. Second, if $1 \leq s < l$, then $w_1 \to w_2 - \ldots - w_{s+1} \leftarrow w_{s+2}$ is a complex in $H_{l-1}$ by (b) for $H_{l-1}$ which is saved in $H_l$ by the previous observation. Third, if $s = l$, then one derives by using (a) for $H_{l-1}$ that $w_r - w_{r+1}$ in $H_{l-1}$ for $r = 2, \ldots, l$, $w_1 \to w_2$ or $w_1 - w_2$ in $H_{l-1}$, $w_{l+1} \leftarrow w_{l+2}$ or $w_{l+1} - w_{l+2}$ in $H_{l-1}$. Evidently no other edge exists in $H_{l-1}$ among $\{w_1, \ldots, w_{l+2}\}$ and Lemma 7.4 (necessity) says $D_M \langle w_1, w_{l+2}| + w_2 \rangle$ and $D_M \langle w_1, w_{l+2}| + w_{l+1} \rangle$. Shortly, the collection of conditions from the item **(iii)** of the algorithm is satisfied and, by the construction of $H_l$, $w_1 \to w_2$ and $w_{l+1} \leftarrow w_{l+2}$ in $H_l$ and hence, by the first mentioned observation, $w_1 \to w_2 - \ldots - w_{l+1} \leftarrow w_{l+2}$ is a complex in $H_l$.

Thus, the last iteration $H_*$ of the algorithm has the same underlying graph as $G$, and, by (a) for $H_*$, $u \to v$ in $H_*$ implies $u \to v$ in $G_0$. This fact says that there exists a complex in $G$ which contains $u \to v$, which implies by (b) for $H_*$ that there is a complex in $H_*$ containing it and therefore $u \to v$ in $H_*$. Thus, $u \to v$ in $H_*$ iff $u \to v$ in $G_0$ and hence $H_* = G_0$. □